\documentclass[letterpaper]{article} 
\usepackage{aaai23}  
\usepackage{times}  
\usepackage{helvet}  
\usepackage{courier}  
\usepackage[hyphens]{url}  
\usepackage{graphicx} 
\usepackage{natbib}  
\usepackage{caption} 
\usepackage[table]{xcolor}
\usepackage{array}
\usepackage{comment}
\usepackage{natbib}
\usepackage{algorithm}
\usepackage{algorithmic}

\usepackage{newfloat}
\usepackage{listings}
\DeclareCaptionStyle{ruled}{labelfont=normalfont,labelsep=colon,strut=off} 
\floatstyle{ruled}
\newfloat{listing}{tb}{lst}{}
\floatname{listing}{Listing}
\title{Med-EASi: Finely Annotated Dataset and Models for Controllable Simplification of Medical Texts}
\author {
    Chandrayee Basu\textsuperscript{\rm 1},
    Rosni Vasu\textsuperscript{\rm 2},
    Michihiro Yasunaga\textsuperscript{\rm 1},
    Qian Yang\textsuperscript{\rm 3}
}
\affiliations {
    \textsuperscript{\rm 1} Stanford University\\
    \textsuperscript{\rm 2} University of Zurich\\
    \textsuperscript{\rm 3} Cornell University\\
    cbasu@stanford.edu, rosni@ifi.uzh.ch, myasu@stanford.edu, qy242@cornell.edu
}

\usepackage{bibentry}

\begin{document}

\maketitle

\begin{abstract}
Automatic medical text simplification can assist providers with patient-friendly communication and make medical texts more accessible, thereby improving health literacy. But curating a quality corpus for this task requires the supervision of medical experts. In this work, we present \textbf{Med-EASi} (\underline{\textbf{Med}}ical dataset for \underline{\textbf{E}}laborative and \underline{\textbf{A}}bstractive \underline{\textbf{Si}}mplification), a uniquely crowdsourced and finely annotated dataset for supervised simplification of short medical texts. Its \textit{expert-layman-AI collaborative} annotations facilitate \textit{controllability} over text simplification by marking four kinds of textual transformations: elaboration, replacement, deletion, and insertion. To learn medical text simplification, we fine-tune T5-large with four different styles of input-output combinations, leading to two control-free and two controllable versions of the model. We add two types of \textit{controllability} into text simplification, by using a multi-angle training approach: \textit{position-aware}, which uses in-place annotated inputs and outputs, and \textit{position-agnostic}, where the model only knows the contents to be edited, but not their positions. Our results show that our fine-grained annotations improve learning compared to the unannotated baseline. Furthermore, \textit{position-aware} control generates better simplification than the \textit{position-agnostic} one. The data and code are available at https://github.com/Chandrayee/CTRL-SIMP.
\end{abstract}

\section{Introduction}\label{subsec:intro}
Health literacy refers to our knowledge, and our ability to obtain, process, and understand health information and services to make appropriate health decisions \cite{literacy2004prescription}. Low health literacy has several adverse effects including poor patient self-care, lack of timely communication of health issues, and even increased risk of hospitalization and mortality \cite{king2010, berkman2011low, tajdar2021low}. Low digital health literacy makes it harder for consumers to disambiguate between reliable medical information \cite{nih, savery2020} and unreliable ones, accelerating the spread of medical misinformation, as in the case of COVID-19 \cite{bin2021covid}. Despite the promise of automated medical text simplification in mitigating this problem, there are limited datasets and open-source libraries for this task \cite{siddharthan2014, phan2021scifive, msd, van2019}. In this work, we present the very first finely annotated dataset \emph{Med-EASi} for medical text simplification and describe two different T5-based models $ctrlSIM$ and $ctrlSIM_{ip}$ that enable controllable simplification.
We define \textit{controllability} as the ability of a user to selectively simplify the contents of a short medical text. Existing unsupervised models \cite{prabhumoye2018, subramanian2018, yelp, asset} and hierarchical tagging and text editing models \cite{malmi2019encode, mallinson2022edit5, mallinson2020felix} can be trained directly on unlabeled datasets like \cite{msd} for medical text simplification. However, we observe the need for fine-grained annotations to enable word or phrase level \textit{controllability} over the model outputs and for elaborative simplification \cite{srikanth2020}. 

\begin{table*}[]
\small
\def\arraystretch{1.2}
\centering
\begin{tabular}{p{8cm} p{8cm}}
\cline{1-2}
\textbf{Expert} & \textbf{Simple} \\ \cline{1-2}
1. People with \textcolor{magenta}{positive fecal occult blood tests} require colonoscopy, as do those with \textcolor{magenta}{lesions} seen during a sigmoidoscopy or an imaging study. & People with \textcolor{magenta}{positive fecal occult blood tests} require a colonoscopy, as do those with \textcolor{magenta}{sores} seen during a sigmoidoscopy or an imaging study. \\
\cline{1-2}
2. During childhood, \textcolor{magenta}{she suffered from partially collapsed lungs} twice, had pneumonia four \textcolor{magenta}{to} five times \textcolor{magenta}{a year, as well as a ruptured} appendix and \textcolor{magenta}{a tonsillar cyst}. & During childhood, \textcolor{magenta}{suffered from collapsed lungs} twice, had pneumonia four \textcolor{magenta}{or} five, \textcolor{magenta}{(ruptured)} appendix and \textcolor{magenta}{a tonsillar cyst}. \\
\cline{1-2}
3. \textcolor{magenta}{They} include a wide variety of pathogens, such as Escherichia Salmonella, Vibrio, Helicobacter, and many other notable genera. & \textcolor{magenta}{These bacteria} include a wide variety of pathogens, such as Escherichia Salmonella, Vibrio, Helicobacter, and many other notable genera. \\
\cline{1-2}
4. \textcolor{blue}{Monozygotic twins have a concordance of about 45 \%}. & \textcolor{blue}{When comparing monozygotic twins, we found that they had a concordance of about 45 \%}. \\
\cline{1-2}
5. \textcolor{blue}{CT also is required to accurately assess skull base bony changes, which are less visible on MRI}. & \textcolor{blue}{Computed tomography (CT) also helps to assess these changes}. \\
\cline{1-2}
\end{tabular}
\caption{Med-EASi is the first finely annotated dataset for controllable medical text simplification. Our T5-based models with multi-angle training: $ctrlSIM$ and $ctrlSIM_{ip}$ allow users to selectively edit contents of the complex medical text by \textit{elaboration}, \textit{replacement}, \textit{deletion} or \textit{insertion}. The above examples are generated by our best model $ctrlSIM_{ip}$. Replacements are shown in magenta and elaborations in blue.} 
\label{tab:model_outputs}
\end{table*}

Crowdsourcing annotations for medical texts is challenging, which explains the dearth of datasets and research in this domain, compared to general text simplification. To obtain high-quality annotations, we must recruit a specific sub-population of workers with domain expertise \cite{pico}. Furthermore, when it comes to evaluating the quality of simplification, only experts can judge the correctness and relevance of the added content. On the other hand, only the laymen audience can validate the readability and comprehensibility of the model outputs. Therefore, we deploy a novel data annotation format for this research, involving both medical experts and layman crowd-workers.

We make the following contributions:
\begin{itemize}
    \item \textbf{Dataset:} We finely annotate two existing parallel medical text simplification corpora with four kinds of textual transformations, viz. elaboration, replacement, deletion, and insertion of new content. 
    \item \textbf{Expert-layman-AI annotation:} We divide the annotation tasks between medical experts and layman crowd-workers depending on the syntactic and domain-specific complexity of the example texts. We assist layman crowd-workers by providing AI-generated annotations.
    \item \textbf{Controllable models:} We train two multi-angle \cite{Tafjord2021Macaw} text simplification models with T5-large backbone. Our models perform controllable simplifications with satisfactory outputs.
\end{itemize}

\section{Related Work}
 Text simplification in the medical domain is mostly restricted to paraphrasing \cite{abrahamsson2014medical}. More recent transformer-based approaches follow developments in general text simplification \cite{zhang2017sentence, nisioi2017exploring, jiang2020neural}. These models treat text simplification as monolingual translation without controllability and deploy tricks like auto-completion \cite{van2020automets} and unlikelihood training \cite{welleck2019neural, devaraj2021} to mitigate for low resources, and common pitfalls of LLMs, e.g. hallucination, and neural text degeneration. 
 
 Meanwhile, research in automatic non-medical text simplification has been burgeoning, with the introduction of large parallel corpora \cite{zhu2010, woodsend2011, coster2011, xu2015, paetzold2017}. The state-of-the-art models can be classified into edit-based \cite{malmi2019encode, mallinson2020felix, agrawal2021non, omelianchuk2021text, cumbicus2021syntax, mallinson2022edit5} and text-to-text models. The creation of multi-references has enabled the models to explicitly learn different kinds of textual transformations, viz. \textit{lexical changes} (e.g. paraphrasing), \textit{syntactic modifications} (e.g. reordering of concepts, splitting texts, reducing sentence length, etc.) and \textit{compression} (e.g. deleting peripheral information irrelevant to the target domain) \cite{asset}. Our approach is similar to the text-to-text models that exert controllability using task-specific prompts \cite{keskar2019ctrl, dathathri2019plug, kariuk2020cut, brown2020language, reif2021recipe, styleptb, xu2022text}. 
 
 However, unlike other models that control for specific attributes of the generated simplification like compression ratio, word rank \cite{martin2019controllable}, level of paraphrasing \cite{maddela2020controllable}, grade-level \cite{nishihara2019controllable} etc., we develop a single end-to-end model that can perform all the edits that simplification entails, similar to edit-based models, while also allowing users to select the content to be edited and the desired form of edit.   

\section{Features of Text Simplification} 
We define text simplification as the process of reducing the linguistic complexity of a text, while still retaining the original information content and meaning \citep{siddharthan2014}. A domain-specific text undergoes various kinds of edits to reach the final simple form. We explore four kinds of textual transformations, defined as follows: 

\emph{Deletion:} removal of any word, phrase, sub-statement, or full sentence from the expert text

\emph{Insertion:}  addition of words, phrases, and sentences that just change the style of the text or fix errors. They do not provide any extra information about any term in the text.

\emph{Replacement:}  replacing any complex word or phrase in the expert text with simpler words or phrases. 
Unlike elaboration, the original term is missing in the simple text.

\emph{Elaboration:} extra information or definitions of original content in the expert text, added as a word or a phrase to the simple text. Typically, the term being elaborated is also present in the simple, optionally replaced by its synonym. We consider these as two distinct types: elaborations where part of the original phrase is preserved (\textit{type 1}) and elaborations where original content is fully replaced by its synonyms (\textit{type 2}).

While elaboration can be broken down into keep, delete, and insert content and replacement can be broken down into delete and insert content \cite{mallinson2020felix}, we treat these transformations differently for more human-interpretable controllability.

\section{Creating Med-EASi}
\subsection{Existing Parallel Corpora}
To create Med-EASi, we leverage existing parallel corpora ([expert $E$, simple $S$] text pairs) for medical text simplification. Based on the evaluation by \citet{basu2021automatic}, we select two publicly available datasets for annotation: SIMPWIKI \cite{van2019} and MSD \cite{msd}. We sorted the text pairs by Levenstein Similarity and Compression ratio \cite{easse} and selected samples for annotation, to cover a wide range of these metrics. 

\begin{figure}[tp]
	\centering
	\includegraphics[width=0.48\textwidth]{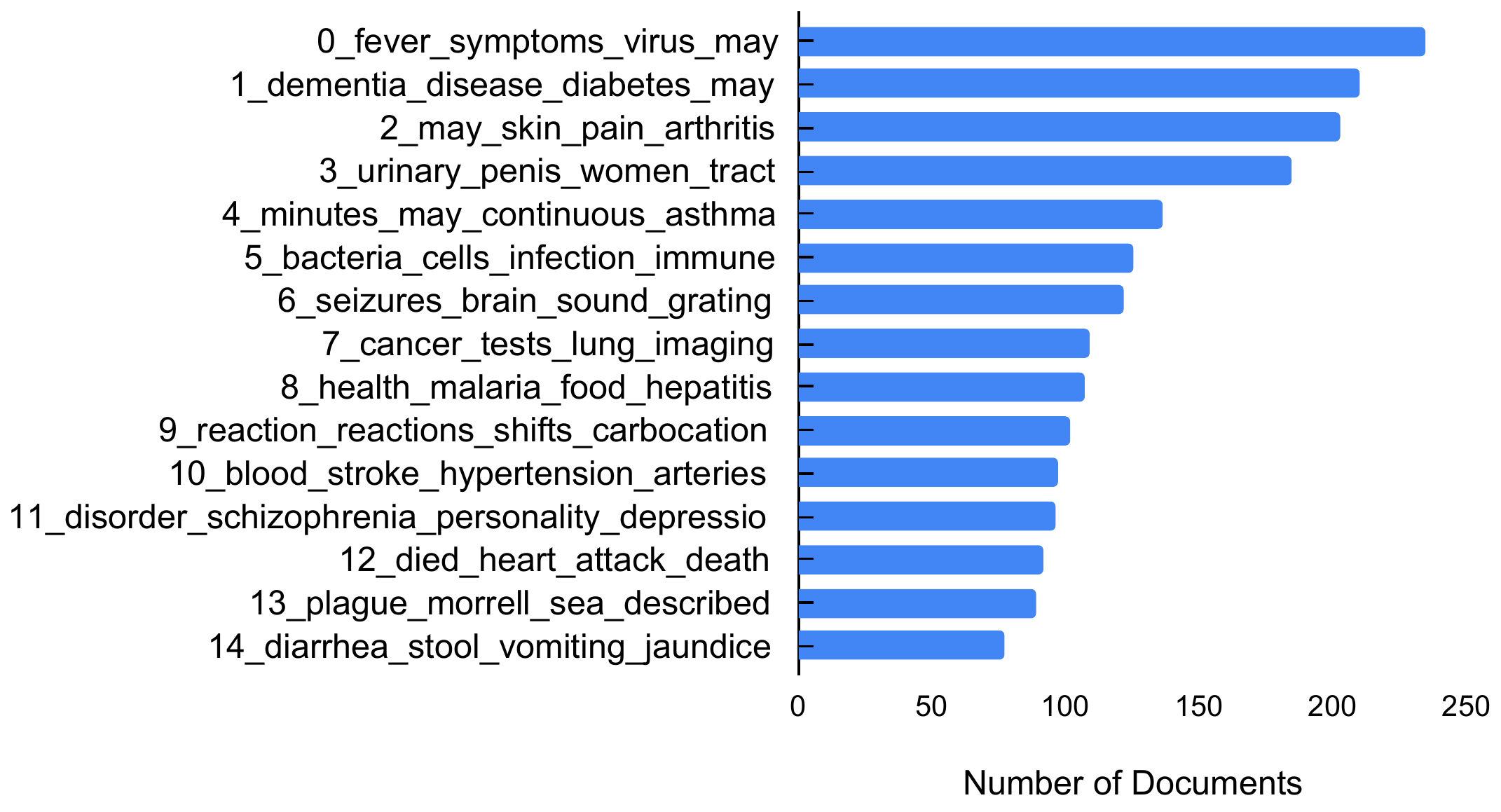}
	\caption{Topic distribution of Med-EASi with top-3 tokens on the x-axis.} 
	\label{fig:topic}
\end{figure}

\subsection{Expert-layman-AI crowdsourcing}
 We recruited experts and layman crowd-workers to annotate Med-EASi with four textual edits (see section Features of Text Simplification), depending on the complexity of the texts. Many SIMPWIKI text pairs have high Levenstein similarity ($>$ 0.7), while MSD examples are more dissimilar and complex. We, therefore, allocated most of SIMPWIKI to the layman crowd-workers and left all of MSD and more difficult SIMPWIKI examples for the medical experts. 
 
 \subsection{Annotations}
Crowdsourcing annotations from layman workers is advantageous due to the sheer number of available annotators. This, however, acts contrary to the whole purpose of medical text simplification: the texts are already inaccessible to layman workers. To make the annotation task easier, we, therefore ask the layman workers to choose between two possible AI-generated annotations. Specifically, we use difflib \cite{difflib} to identify how text A was transformed into text B using four types of edits: replace, keep (equal), delete and insert. We post-processed difflib's output to generate annotation as \textit{Uterine cancer, also known as womb cancer, is any type of cancer that $<$rep$>$emerges$<$by$>$starts$<$/rep$>$ from the tissue of the uterus.}, where \textit{$<$rep$>$} stands for replace.

We used a SpanBERT-based coreference resolution model \cite{joshi2019bert} to identify spans in the simple text that referred to some entity in the expert text. We passed concatenated expert and simple texts to this model. We marked a span in the simple text as elaboration if it was longer than the reference span in the expert text. Resultant annotation looked like this: \textit{In his $<$elab$>$Nobel lecture$<$by$>$Nobel Prize lecture$<$/elab$>$, $<$elab$>$Lewis$<$by$>$E.B. Lewis$<$/elab$>$ said "Ultimately, comparisons of the (control complexes) throughout the animal kingdom should provide a picture of how the organisms, as well as the (control genes) have evolved."}, where \textit{$<$elab$>$} stands for elaboration. 

\subsection{Layman crowdsourcing}
We recruited layman crowd-workers who were fluent in English from Toloka \cite{toloka} and followed a three-step training protocol. In step 1, we asked crowd-workers to watch a video describing the purpose of the project, our dataset, and the annotation tasks. The video was followed by 2 experience-related questions and 1 exam question. We asked workers to indicate the frequency of their interaction with medical data in English (messages with doctors, medical bills, Wikipedia, etc.), and their level of experience in NLP tasks. In the exam question, we asked them to select the correct annotation format for replacement. Workers who had watched 85 \% of the video, had experience interpreting medical data in English, and were able to answer the exam question correctly were selected for the second round of training (83 \% acceptance rate).

In step 2, we assigned 12 tasks. Each task contained an expert-simple text pair and two possible aforementioned annotations. We asked the workers to choose the correct annotation from the two options or select None of the above. Workers with $>$ 80 \% correct answers were automatically selected for the next stage of training (80 \% acceptance rate). 

In step 3, all the tasks selected required corrections from the workers. These were examples where our automatic annotation schemes had failed. We manually checked the correct annotations and accepted any annotation that had at most one missing annotation (e.g. missed one replacement) and high overlap with ground truth spans. Workers with $>$ 70 \% accepted answers were selected for the actual annotation task (15 \% acceptance rate).  

The final annotation task design matched that of training step 2. We offered a bonus for writing the correct annotation. We got 3 annotations for each data pair from 3 workers. We aggregated the annotations using Dawid-Swene aggregation method \cite{dawid1979maximum} and automatically accepted the ones with at least 90\% confidence. We passed the ones with lower confidence to an expert. 

\subsection{Expert crowdsourcing}
We recruited 5 experts from Upwork: two medical doctors, two medical students, and one biomedical research scientist, all with experience in data annotation for NLP. We asked the experts to watch the full instruction video and gave them 20 annotation tasks. The annotations were checked by two of our in-house team members. We relied on their medical knowledge significantly and clarified any disagreement with further discussion. All 5 experts preferred to provide correct annotations directly without referring to AI-generated ones.

\subsection{Dataset statistics}
We annotated all of MSD, because of its clinical nature and diversity of transformations and approximately 1500 text pairs from SIMPWIKI. The resulting Med-EASi dataset contains a total of 1979 expert-simple text pairs. 

\noindent
\textbf{Topic distribution of the dataset: }To understand the domain-complexity of the data, we identified the medical concepts in each text pair and their Unified Medical Language Systems (UMLS) \cite{bodenreider2004unified} representations using QuickUMLS \cite{soldaini2016quickumls}. Med-EASi covers a total of 3909, 3304 unique medical concepts in the expert and simple texts respectively, and a total of 4478 concepts across all text pairs. The topic distribution computed using BERTopic \cite{grootendorst2022bertopic} shows wide coverage of medical subdomains like infectious diseases, cardiology, neurology, etc.  (Figure \ref{fig:topic}). 

\noindent
\textbf{Quality: }Following \citet{basu2021automatic} we measure data diversity using reference-less quality metrics from EASSE library \cite{martin2018reference}. The Levenstein similarity ($0.689\pm0.22$), the fraction of words added ($9.347\pm11.57$), deleted ($10.574\pm12.38$), kept ($12.792\pm10.01$), and the compression ratio ($1.025\pm0.59$) were computed. Med-EASi has overall acceptability $98.989\%$ for expert text and $98.737\%$ for layman texts computed using COLA-trained DistilBERT classifier \cite{morris2020textattack, warstadt2019neural}. The readability is estimated using the Flesch Kincaid readability grade (FKGL), measured as the minimum education level required to read and understand a text, expressed as an empirical function of total words, total sentences, and total syllables \cite{kincaid1975derivation}. The expert texts ($12.47\pm5.28$) have statistically significant differences in the readability grade (paired t-test with $p<0.001$) compared to the layman text ($10.491\pm4.98$).

\section{Multi-Angle Controllable Simplification Model}
Any model performing simplification  must have three capabilities: 1. predicting the span of the expert text that must be altered, 2. predicting the alteration or operation on each span, and 3. predicting the additional (in case of elaboration) or alternative (for replacement) contents. In \emph{Controllable text simplification}, a user should be able to indicate the complex contents that must be deleted, replaced, or elaborated. In such cases, instead of asking the model to predict the span that must be transformed, the user can provide the span as input and the model should be able to incorporate the requested transformation into the generated simplification. Following our definition of \textit{controllability} \ref{subsec:intro}, we aim to develop a model that can simplify short medical texts, both with and without controllability instructions. 

Med-EASi contains diverse training pairs, each with a different set of textual transformations. Therefore, we utilize the flexible nature of seq-2-seq models \cite{raffel2020exploring} and finetune T5-large with a combination of left-to-right generation and infilling. We use variable task descriptions to accommodate heterogeneous forms of inputs and outputs, similar to MACAW \cite{Tafjord2021Macaw}. This approach is called multi-angle training, where each component of the input and the output is a \textbf{slot} and the input-output combination is called an \textbf{angle}. Some examples of slots include expert text, phrases to be elaborated, or content that must be replaced. 

\subsection{Slots and Angles}
We consider each example as a set of slots $S_i$ and corresponding values $V_i$. Angle $A_i = S_{Si} \rightarrow S_{Ti}$ is a combination of source or input slots $S_{Si}$ and target or output slots $S_{Ti}$. The slots are a way to decompose the task descriptions and prompts into human-interpretable instructions. It also allows easy recovery of the final simplification from the generated output with standard post-processing. The requested output slots are concatenated as task descriptions at the beginning with no associated values. We keep the \texttt{\small \$simple\$} slot as the last requested slot. We hoped that the generated simplification will be more accurate if conditioned on the predicted edit spans. While identifying angles, we use uppercase abbreviations of the slot names: We identify the slots with uppercase letters, D: deletion, I: insertion, R: replacement, X: elaboration, E: expert text, S: simple text, Ea: annotated expert text, Sa: annotated simple text. See Table. \ref{tab:model_inputs} for color-coded examples of slots and angles.

\subsection{Model versions} \label{subsec:model_versions}
We train one baseline model without annotated data. We hypothesize that controllability is not only desirable but can also improve text-to-text model performance by providing additional supervision signals. To test this, we develop four versions of T5-large: two control-free and, two controllable versions. \\

The \textbf{control-free} versions are referred to as $SIM$ and $SIM_{ip}$ respectively ($ip$ stands for in-place annotation). The former predicts all the spans that must be transformed and the corresponding transformations. This version of the model also learns to predict what transformations are invalid for a given example. The latter directly generates the annotated simplification, from which the simple text is extracted with post-processing. \\

Like the control-free models, in case of the \textbf{controllable models} too, we experiment with two different input-output formats: \textit{position-agnostic} and \textit{position-aware}. We refer to the controllable models as $ctrlSIM$ and $ctrlSIM_{ip}$ respectively. $ctrlSIM$ allows users to input the words or phrases they expect to be edited by replacement or elaboration, while $ctrlSIM_{ip}$ requires users to highlight the same contents in place. 

\begin{table*}[]
\small
\def\arraystretch{1.2}
\begin{tabular}{p{1.5cm}p{1.5cm}p{5.5cm}p{7.5cm}}
\cline{1-4}
\centering
\textbf{model} & \textbf{angle} & \textbf{input} & \textbf{output}   \\ \cline{1-4} 
    $ctrlSIM$ &  ERi$\rightarrow$RS & \textcolor{red}{\$replace\$ ; \$simple\$ ;} \textcolor{blue}{\$expert\$ = }Ankles, knees, elbows, and wrists are usually involved. ; \textcolor{blue}{\$replace\_in\$ =} [involved]   &   \textcolor{teal}{\$replace\$ = }[involved $<$by$>$ affected] ; \textcolor{teal}{\$simple\$ = }Ankles, knees, elbows, and wrists are usually affected.            \\ \cline{1-4}
   $ctrlSIM_{ip}$   &   Ea$\rightarrow$Sa &   \textcolor{red}{\$annotated\textunderscore simple\$ ;} \textcolor{blue}{\$annotated\textunderscore expert\$ = }$<$elab$>$Allergic bronchopulmonary aspergillosis,$<$extra\textunderscore id\textunderscore 0$>$ $<$rep$>$a hypersensitivity reaction to Aspergillus species that$<$extra\textunderscore id\textunderscore 1$>$ occurs most commonly in people with asthma.        &     \textcolor{teal}{\$annotated\textunderscore simple\$ =} $<$elab$>$Allergic bronchopulmonary aspergillosis,$<$by$>$Allergic bronchopulmonary aspergillosis, which affects the larger airways, can cause mucus plugs that block the airways and lead to bronchiectasis.$<$/elab$>$ $<$rep$>$a hypersensitivity reaction to Aspergillus species that$<$by$>$It is an allergic reaction to the fungus Aspergillus and$<$/rep$>$ occurs most commonly in people with asthma.     \\ \cline{1-4}
\end{tabular}
\caption{The table shows two forms of instructions for controllability, $ctrlSIM$: contents of the expert text that must be edited without reference to their position in the expert text and $ctrlSIM_{ip}$: expert text with in-place annotation of spans to be edited and the desired edit types. The task descriptions are color-coded with red, the input slot names with blue, and the output slot names with teal. The values corresponding to the slots are in regular text color.}
\label{tab:model_inputs}
\end{table*}

\section{Training}
\subsection{Data, Angles, and Slots}
We allocated 75 \% or 1398 data points for training and set aside 10 \% or 197 data points as dev set for hyperparameter tuning. The dev set matched the data distribution of the test set with some unseen UMLS terms. In our training data, we have a total of 445 elaborations, 2044 replacements, 512 insertions, and 905 deletions. So approximately 50\% are replacement operations. 

$SIM$ and $SIM_{ip}$ are trained on fixed angles: E$\rightarrow$RXDIS and E$\rightarrow$Sa respectively. In the case of examples with missing slots, all the empty slots contain the same token \texttt{\small $<$extra\textunderscore id\textunderscore 0$>$} as value. While the input-output format for $ctrlSIM$ is similar to $SIM$, the model here only outputs slots present in the example and is not required to know which slots are empty. Table \ref{tab:model selection} shows all the training angles of $ctrlSIM$. $ctrlSIM_{ip}$ is trained on two different angles $E\rightarrow Sa$ and $Ea\rightarrow Sa$. Annotated expert text marks the beginning of the content to be transformed with the corresponding edit type tag (\texttt{\small$<$rep$>$} or \texttt{\small$<$elab$>$}) and adds a special token at the end of the content. The token is different for elaboration and replacement. We do not expect the user to know what content must be inserted or deleted and leave that to the model to figure out.

\subsection{Pretrained Models and Hardware}
We experimented with two different pretrained models as backbones for our text simplification models: T5-large \cite{raffel2020exploring} and SciFive-large(+pubmed+pmc) \cite{phan2021scifive}. While T5 is trained on 750 GB of web-crawled data, i.e. Colossal Clean Crawled Corpus (C4) \cite{raffel2020exploring}, it is not specifically fine-tuned for medical text generation. SciFive, on the other hand, is T5 retrained on combinations of C4, PubMed database of 32 million citations and abstracts \cite{pubmed}, and PubMed Central (PMC), a corpus of free full-text articles in
the domain of biomedical and life sciences \cite{pmc}. We trained our models on two different servers, one with two GeForce RTX 3080 and another with three GeForce RTX 2080 Ti.

\subsection{Decoding}
Greedy decoding worked well for our output generation. We tested several beam sizes for beam search. A beam size of 10 handled the open-ended nature of elaboration and replacement tasks adequately, where the model is expected to leverage the knowledge of the pretrained backbone, beyond our training corpus. Nucleus sampling did not improve SARI score (next section) much and brought down the other scores.

\subsection{Metrics}
We evaluate the quality of simplification using SARI: a well-known metric that measures the goodness of words added, deleted, and kept by the simplification model, by comparing the generated text against the input and multiple references \cite{xu2016optimizing}. 

We report the results of our two controllable models in terms of SARI and its sub-scores ADD, DEL, and KEEP, wherever applicable, and individually for each type of output slot. We considered several possible errors in our models including missing slots or predicting undesired slots. Although rare, we had a few examples with such errors. While computing SARI-based metrics we only considered examples with slots that have values in the true label as well as generated output. We ignored the following: examples that have slots with no values in the true label or in the generated output, examples that have slots with values in the true label but no values in the output, or vice versa.

Slot-wise we computed SARI for generated simplification, ADD score for insertion, KEEP and ADD for elaboration type 1 and, ADD and DEL for replacement. In order to predict whether $ctrlSIM$ is able to identify the content that must be elaborated or replaced and the corresponding valid additions, we split the output for these two slots into two parts: 1. textual span that must be edited ($S_{pre}$), and 2. the resulting span after the transformation ($S_{post}$), separated by \texttt{\small$<$by$>$} in model inputs and outputs. SARI DEL can be applied to $S_{post}$ to evaluate whether certain content is present in the expert text, but not in the simple text. It, however, cannot evaluate if a model is able to predict the content to be deleted correctly. The same problem exists in the case of $S_{pre}$ for replacement. To account for the deleted span of the expert text we modified the DEL score into ALTDEL score. Following notations of \citet{xu2015} (ignoring the n-grams $g$ for brevity), ALTDEL score can be written as:

\begin{equation}
    p_{altdel} = \frac{(I \cap O) \cap \bar{R}}{O}
\end{equation}

\begin{equation}
    r_{altdel} = \frac{(I \cap O) \cap \bar{R}}{(I \cap \bar{R})}
\end{equation}

where $p$ and $r$ refer to precision and recall respectively. The predicted span (O) to be deleted must be present in the expert text (I) and absent from the simple text (R). Please note that for the rest of the paper, we revert back to using I and R for insertion and replacement respectively.

We also evaluate our controllable models' performance using ROUGE-L \cite{lin2004rouge} for recall. Since the ultimate goal is to generate simplifications that are accessible to laymen with different levels of medical knowledge, we report their readability using FKGL.

\begin{table*}[]
    \centering
    \begin{tabular}{lllccccccc}
        \hline
         models & base & angle & SARI & ADD & DEL & KEEP & FKGL \\
         \hline
         Baseline & T5 & E$\rightarrow$S &  39.14 & 9.34 & 67.8 & 40.29 & 11.1 \\
         $SIM$ & T5 & E$\rightarrow$RXDIS & 36.47 & 6.9 & 62.25 & 40.25 & 10.05 \\
         $SIM_{ip}$ & T5 & E$\rightarrow$Sa & 37.61 & 7.15 & 64.65 & 7.15 & 11.87 \\
         $ctrlSIM$ & T5 & $multi$ & 39.28 & 7.13 & 66.94 & 43.78 & 11.04 \\
         $ctrlSIM$ & Sci5 & $multi$ & 38.07 & 7.53 & 65.39 & 41.3 & 11.28 \\
         $ctrlSIM_{ip}$ & T5 & $multi_{ip}$ & 40.89 & 6.58 & 75.16 & 40.94 & 11.41 \\
         $ctrlSIM_{ip}$ & Sci5 & $multi_{ip}$ & 39.42 & 4.89  & 73.6 & 39.78 & 10.87 \\
         \hline
    \end{tabular}
    \caption{Dev set performances for various model versions, $multi$: [E$\rightarrow$S, E$\rightarrow$DIS, ERi$\rightarrow$DRS, ED$\rightarrow$IS, EDXi$\rightarrow$XS, ERi$\rightarrow$RS, ERiXi$\rightarrow$DRXS, E$\rightarrow$DS, EXi$\rightarrow$XS, ERiXi$\rightarrow$RXS, EDRi$\rightarrow$RS, EDRiXi$\rightarrow$RXS, E$\rightarrow$IS, ED$\rightarrow$S, EXi$\rightarrow$DXS], $multi_{ip}$:[E$\rightarrow$Sa, Ea$\rightarrow$Sa]}
    \label{tab:model selection}
\end{table*}

\subsection{Hyperparameter tuning}
We fine-tuned the baseline, as well as all 4 versions of our models for 30 epochs, each with batch sizes 4, 8, 16, 32, and 64, and a constant learning rate of 8e-06. \\

Since the primary goal of our model is to output correct simplification, we use SARI instead of dev set error, for model selection. When computing SARI for each dev example, we ignored the examples where the model was unable to output all the requested slots. We observed that while dev set error goes up with more training, the model is able to understand the meaning of the slots better as well as able to add simpler words with further training. 
While KEEP, DEL, and SARI did not change much beyond epoch 15, ADD score and FKGL improved with further training.

\begin{table*}[]
    \centering
    \begin{tabular}{lllcccccccccc}
        \hline
         models & base & angle & SARI & ADD & DEL & KEEP & FKGL & ROUGE-l\\
         \hline
         $ctrlSIM$ & T5 & $multi$ & 39.63 & 8.99 & 70.67 & 39.2 & 10.55 & 0.41\\
         $ctrlSIM_{ip}$ & T5 &  
         $multi_{ip}$ & 40.2 & 8.51 & 70.07 & 42.04 & 11.09 & 0.43\\
         \hline
    \end{tabular}
    \caption{Test set performances for various model versions, $multi$: [E$\rightarrow$S, E$\rightarrow$DIS, ERi$\rightarrow$DRS, ED$\rightarrow$IS, EDXi$\rightarrow$XS, ERi$\rightarrow$RS, ERiXi$\rightarrow$DRXS, E$\rightarrow$DS, EXi$\rightarrow$XS, ERiXi$\rightarrow$RXS, EDRi$\rightarrow$RS, EDRiXi$\rightarrow$RXS, E$\rightarrow$IS, ED$\rightarrow$S, EXi$\rightarrow$DXS], $multi_{ip}$:[E$\rightarrow$Sa, Ea$\rightarrow$Sa]}
    \label{tab:test results}
\end{table*}

\section{Results}
We allocated 15 \% of Med-EASi or 300 text pairs for evaluation. We sampled the test set to cover wide ranges of Levenstein Similarity and representative UMLS terms. To understand whether our model is able to retrieve its pretraining knowledge in the text simplification task, we kept 34 \% of the 300 text pairs with 672 previously unseen UMLS terms. We sampled the rest 200 pairs at random from the original data distribution. Based on dev set performance we picked the top model checkpoints and evaluated their performances on the above test set.

We test the following features of both $ctrlSIM$ and $ctrlSIM_{ip}$ with our test results. 

\subsection{Overall quality of  simplification}

We trained the best model $ctrlSIM_{ip}$ once again, for 60 epochs, with 10 \% warm-up steps and a cosine learning rate decay from epoch 27 onwards. The new dev set performance showed $41.36$ SARI score and $9.05$ ADD score, an improvement over the prior version. The model achieves this result with a batch size of 4 at epoch 13. 

Overall, we obtained SARI scores comparable with the performances of the top text simplification models on other datasets \cite{mallinson2020felix}.

\subsection{Controllability}

We test if the controllable models are able to perform the task requested by the users. To do so, we split the test results of the two models by angles.

\begin{figure}[tp]
	\centering
	\includegraphics[width=0.4\textwidth]{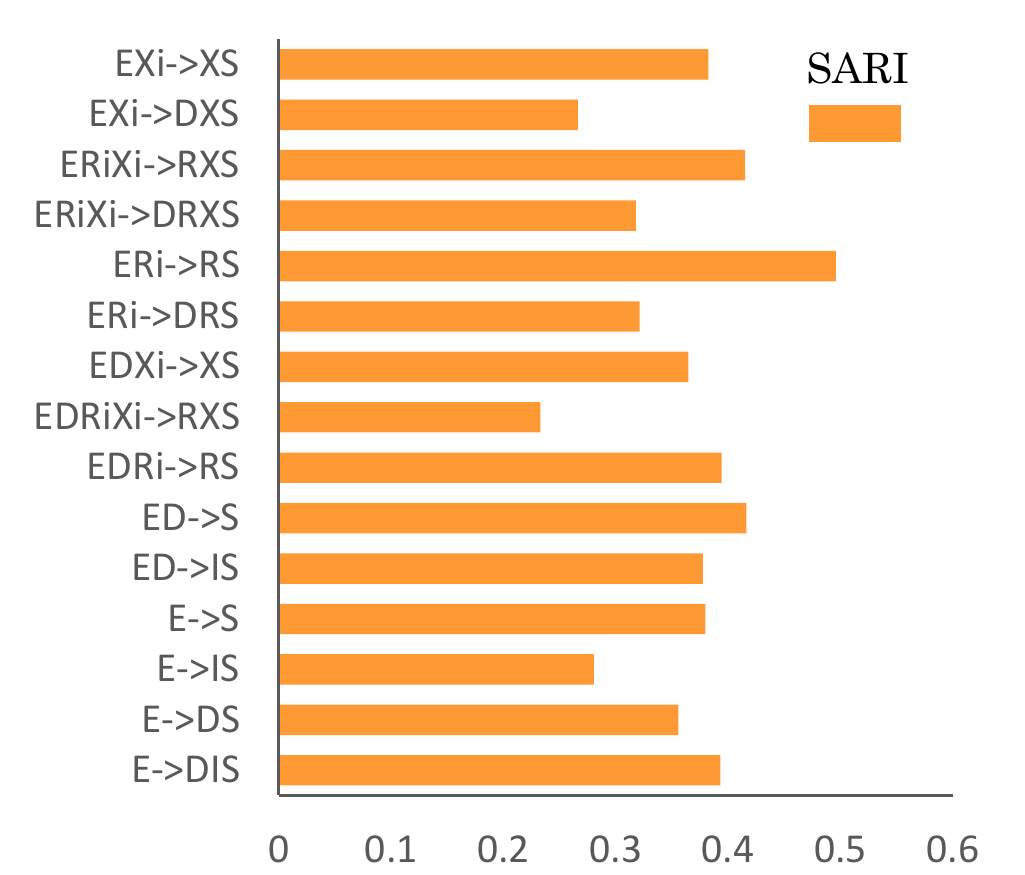}
	\caption{SARI scores of $ctrlSIM's$ outputs arranged by angles}
	\label{fig:sari by angle}
\end{figure}

Figure \ref{fig:sari by angle} shows the average SARI score of the simplifications generated by $ctrlSIM$. We observe that the angle $ERi\rightarrow RS$ (when the user provides the content that must be replaced, $Ri$ in addition to the expert text $E$) achieves the highest score of $0.49$. Such a high score could be attributed to the large number of instances of this angle in the training data. Furthermore, when the model predicts other slots prior to the simplification, like replacement or elaborations, it improves the quality of the generated simplification, as seen for angles $EXi\rightarrow XS$, $ERiXi\rightarrow RXS$ and $ERi\rightarrow RS$. Likewise, instructing the model to remove certain content results in a simplification with the highest DEL score.

$ctrlSIM_{ip}$ demonstrates the above phenomenon more clearly. The average SARI scores of the simplifications vary significantly by the angles, $0.22$ for $E\rightarrow Sa$ and $0.46$ for $Ea\rightarrow Sa$. 

\subsection{Elaboration and Replacement}

We ask: How are the models performing in open-ended and potentially more complicated generation tasks like elaboration and longer replacements?

We split the test results of the two models by slots. For replacement, we report the average ADD score of the generated replacements. We measure the elaboration quality by the mean of ADD score and KEEP score and call it the elaboration score. $ctrlSIM$ on test data produces an average ADD score of $13.24$ for replacement, an ADD score of $9.85$, a KEEP score of $23.63$, and an overall elaboration score of $16.74$ across all the data points. Overall, we observed that $ctrlSIM$ is able to detect the contents to be deleted much better than addition with an average ALTDEL score of $39.27$. Our low ADD score of the simplified texts, on an average and slot-wise, suggest that the model is unable to add content. One potential solution would be to concatenate additional context to the examples as another input slot. We have such a provision in our existing models that researchers can use in future work. 

$ctrlSIM_{ip}$ performed better deletion and replacement than $ctrlSIM$ with ALTDEL score of $51.96$ for deletion and average ADD score of $18.08$ for replacement. The model displayed similar elaboration skill as $ctrlSIM$ with an overall elaboration score of $16.53$. Note that these results are only from the test examples with controllable angle $Ea\rightarrow Sa$.

Qualitatively, we observe that many times $ctrlSIM_ip$ replaces a text with the correct alternative, and other times fails to perform any replacement. Likewise, when asked to elaborate, sometimes $ctrlSIM_{ip}$ is able to retrieve full forms of medical abbreviations from its prior knowledge, but at other times elaboration is just treated as a change of style (see Table \ref{tab:model_outputs}). Since some of these errors can only be obvious to humans, we resort to human evaluation, in particular expert evaluation.

\subsection{Human Evaluation}
We recruited three medical/biomedical experts to evaluate the quality of $ctrlSIM_{ip}$'s outputs. We also hired two layman native speakers to rate fluency and grammatical correctness. We sampled 50 outputs randomly from our test data for this purpose.\\
For fluency, we used a 4-point scale and a yes/no for grammaticality. We specifically asked our experts to judge the output annotation quality by answering 4 questions: one yes/no question and the remaining 3 to be rated on a 4-point scale (0-3). We asked,

\begin{itemize}
    \item Did the model perform what the user asked for? 
    \item Do the replacements in the output annotation match the replaced contents?
    \item Are the elaborations in the output annotation relevant to the content elaborated?
    \item Are the elaborations satisfactory?
\end{itemize}

\textbf{Fluency:} There is a moderate agreement between the 2 experts ($tau = 0.397$, $p = 0.0018$). The mean scores of $2.32$ and $2.22$ indicate that the generated text is mostly fluent. \textbf{Grammaticality:} Both experts agreed that $40\%$ of the model output is grammatically correct. \textbf{Output annotation Quality:} According to the experts, $22\%$ of the time, the model performed as expected. The mean scores from each expert ($2.36$, $1.86$, and $2.04$) indicate that model-generated content \textit{mostly matches} the replaced content. Furthermore, the added content for elaboration is \textit{mostly relevant} to the content elaborated (mean scores $2.33$, $2.0$, and $2.47$), however, the elaborations were only \textit{somewhat satisfactory} (mean scores $2.27$, $1.2$, and $1.73$).

\section{Summary}
We are constantly fed with medical information online through news articles, popular science magazines, and social media posts. Despite volumes of medical articles being generated every day, low health literacy remains a challenge in healthcare. To make the medical texts more accessible, we create a finely annotated dataset Med-EASi for medical text simplification. Med-EASi consists of several pairs of expert medical texts and their annotated simplifications. It covers a wide range of medical topics and textual complexities and is annotated with four kinds of textual transformations: deletion, insertion, elaboration, and replacement. \\
We leverage the power and flexibility of large LMs like T5 to enable controllable text simplification, where the user can instruct the model to selectively simplify contents of a short medical text. We test two different kinds of controllability, one where the user inputs the content they want to be altered by a specific type of edit, and another, where the user can highlight in-place the same content and mark it with the desired transformation. Both of our controllable models perform at par with the top text simplification models. Our in-place controllable model displays promising results, generating mostly fluent and correct texts. The model is able to replace complex medical content appropriately. 

Overall, we hope that Med-EASi can foster open research in AI-assisted medical text simplification. One of the potential future directions would be to improve the model's elaboration skills by supplying additional contexts like definitions and descriptions of complex medical terms. 

\section*{Acknowledgements}
We are grateful to Bhavana Dalvi Mishra (Senior Research Scientist at Allen Institute for Artificial Intelligence) for her invaluable guidance throughout the duration of this project. We would also like to thank the anonymous reviewers for their insightful feedback. 

We acknowledge Toloka AI for partially funding our crowdsourcing tasks. This work has been partially supported by the Swiss National Science Foundation (SNSF) under contract number 200020\_184994 (CrowdAlytics Research Project).

\appendix

\bibliography{aaai23.bib}

\end{document}